\documentclass[letterpaper]{article} 
\usepackage[submission]{aaai25}  
\usepackage{times}  
\usepackage{helvet}  
\usepackage{courier}  
\usepackage[hyphens]{url}  
\usepackage{graphicx} 
\usepackage{natbib}  
\usepackage{caption} 
\usepackage{algorithm}
\usepackage{algorithmic}
\usepackage{booktabs}
\usepackage{array}
\usepackage{tabularx}
\usepackage{amsfonts}
\usepackage{colortbl}
\usepackage{xcolor}
\usepackage{newfloat}
\usepackage{listings}
\DeclareCaptionStyle{ruled}{labelfont=normalfont,labelsep=colon,strut=off} 
\floatstyle{ruled}
\newfloat{listing}{tb}{lst}{}
\floatname{listing}{Listing}

\makeatletter
\def\showauthors@on{T}
\makeatother
\title{TARDiS : Text Augmentation for Refining Diversity and Separability}
\author{
    Kyungmin Kim,
    SangHun Im,
    GiBaeg Kim,
    Heung-Seon Oh,
}

\affiliations{
    School of Computer Science and Engineering,\\
    Korea University of Technology and Education (KOREATECH)\\
    \{dukes123,tkrhkshdqn, fk0214, ohhs\}@koreatech.ac.kr
}

\begin{document}

\maketitle

\begin{abstract}
Text augmentation (TA) is a critical technique for text classification, especially in few-shot settings. This paper introduces a novel LLM-based TA method, TARDiS, to address challenges inherent in the generation and alignment stages of two-stage TA methods. For the generation stage, we propose two generation processes, SEG and CEG, incorporating multiple class-specific prompts to enhance diversity and separability. For the alignment stage, we introduce a class adaptation (CA) method to ensure that generated examples align with their target classes through verification and modification. Experimental results demonstrate TARDiS's effectiveness, outperforming state-of-the-art LLM-based TA methods in various few-shot text classification tasks. An in-depth analysis confirms the detailed behaviors at each stage.

\end{abstract}

%

\section{Introduction}

Text augmentation (TA) is a critical technique for text classification, especially in few-shot settings where the original data is extremely limited. Incorporating class-specific features during the TA process is crucial to overcoming the limited knowledge derived from few-shot data \cite{anaby2020not,Guo2020,malandrakis2019controlled,sennrich2016improving,wei2019eda,wu2019conditional}. TA leveraging Large Language Models (LLMs) \cite{sahu2023promptmix,sahu2022data,lin2023selective,dai2023auggpt} is notably effective due to the extensive intrinsic knowledge within LLMs. Previous LLM-based TA methods can be generalized into two stages: generation and alignment. In the generation stage, novel examples for a target class are generated using original data. Subsequently, misaligned examples corresponding to incorrect or out-of-distribution (OOD) classes are addressed in the alignment stage.

Existing two-stage TA methods have limitations at each stage. The generation stage typically depends on a single fixed prompt to generate new examples based on seed data. This approach restricts LLMs' inherent knowledge usage and diversity, resulting in two critical limitations: insufficient class-specific features and classification properties. The former can be addressed by employing manual class descriptions, but human intervention is essential and not always feasible or scalable. For the latter, existing methods often focus on a single aspect,  such as either intra-class diversity \cite{sahu2022data,lin2023selective} or inter-class separability \cite{sahu2023promptmix}. In the alignment stage, few-shot settings have inherent weaknesses due to insufficient training data for verifying misaligned examples. This often results in false negatives (FNs), where aligned examples are incorrectly identified as misaligned, and an inability to handle out-of-distribution (OOD) examples. 

To address these limitations, this paper proposes a novel LLM-based TA method, TARDiS (\textbf{T}ext \textbf{A}ugmentation for \textbf{R}efining \textbf{Di}versity and \textbf{S}eparability). In the generation stage, TARDiS uses `spark thoughts', ideas that activate the LLMs' inherent knowledge for each class, enhancing the traditional single-prompt approach with multiple class-specific prompts. To tackle both intra-class diversity and inter-class separability, we present two generation processes employing the multiple class-specific prompts: Semantic Enrichment Generation (SEG) and Contrastive Enrichment Generation (CEG). As illustrated in Figure 1-(b), SEG uses spark thoughts generated from examples within the target class to capture diversity within the target class, whereas CEG uses those generated from both the target class and an ambiguous class, which could be confused with the target class, to improve separability from non-target classes.

\begin{figure*}[h]
    \centering
    \includegraphics[width=0.9\textwidth]{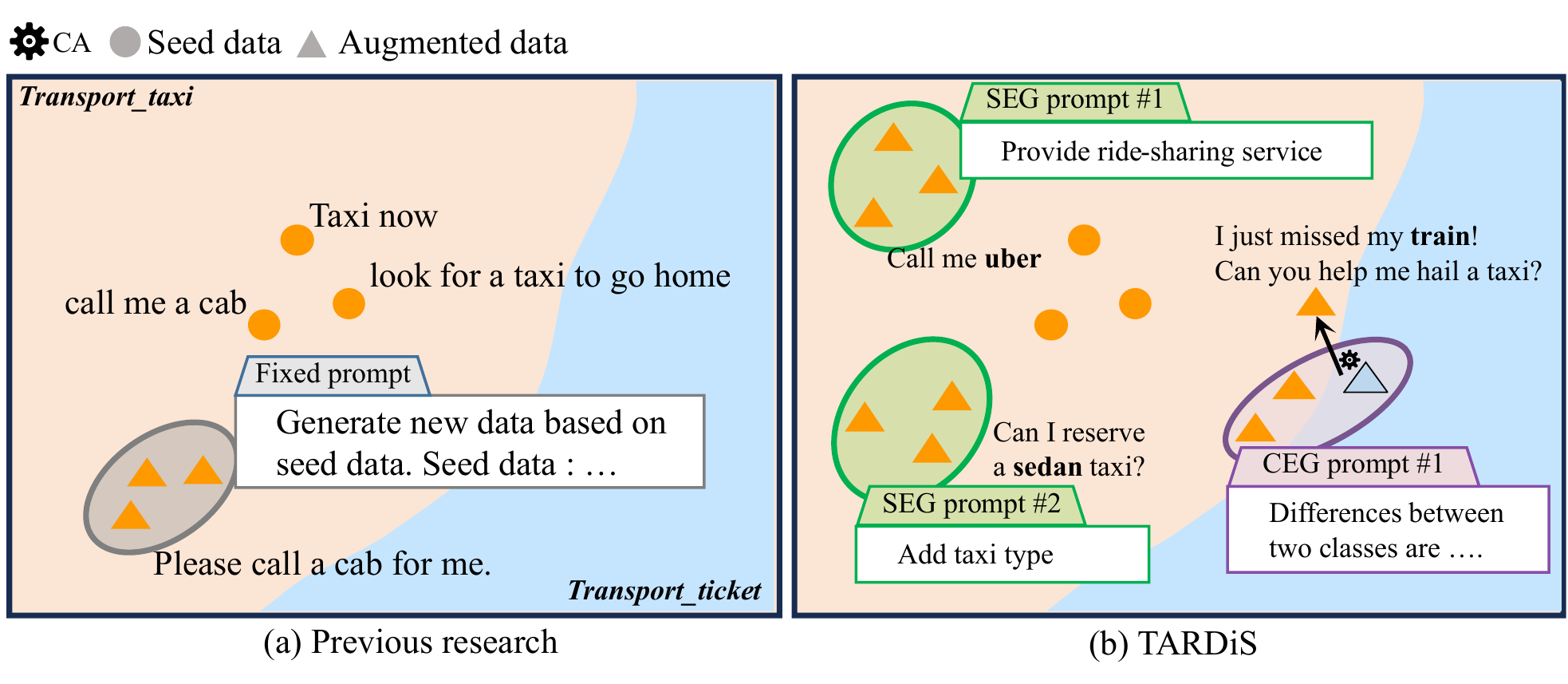}
    \caption{A rectangle denotes a holistic data distribution for \textit{Transport\_taxi} (top left) and \textit{Transport\_ticket} (bottom right) classes, while circles and triangles denote seed data and augmented examples, respectively. (a) Augmented examples from previous research \cite{lin2023selective,sahu2022data}, which generates examples similar to seed data using a single fixed prompt. (b) Augmented examples from TARDiS. SEG and CEG generate various examples enhancing diversity and separability through multiple prompts by spark thoughts. CA aligns misaligned examples with the target classes through verification and modification.}
    \label{fig:fig1}
\end{figure*}

To address the limitations of the alignment stage, we propose a Class Adaptation (CA) method that modifies generated examples to align with the corresponding target class using an LLM  instead of simply relabeling them. Consequently, CA can effectively deal with examples that are misaligned, OOD, or FNs.

The contributions of this paper are summarized as follows:

1.	We propose TARDiS to address the challenges of two-stage TA methods. For the generation stage, we introduce SEG and CEG based on multiple class-specific prompts to enhance diversity and separability, respectively. For the alignment stage, we introduce a CA method to ensure that generated examples align with the corresponding target class through verification and modification.

2.	We demonstrate the effectiveness of TARDiS by achieving SOTA performance on various few-shot text classification benchmarks and investigate detailed behaviors at each stage through an in-depth analysis.

\section{Related Work}
\subsection{Text Augmentation}
TA has been widely studied to enhance the generalization capability of models by generating new examples from seed data. Traditional methods like EDA \cite{wei2019eda} and back translation \cite{sennrich2016improving} create new patterns by altering linguistic characteristics but face limitations in introducing completely new features. Language Model (LM)-based TA methods generate novel examples by leveraging sentence structures \cite{Guo2020,Kim2021,kobayashi2018contextual,wu2019conditional} or modifying parts of the seed data \cite{anaby2020not,Kumar2020} to utilize the knowledge within pre-trained LMs.

Recent advancements in LLMs, such as GPT-3 \cite{Brown2020} and Llama \cite{Hugo2023}, have enabled TA methods to generate novel examples by leveraging the extensive intrinsic knowledge within LLMs. LLM-based TA methods can be generalized into two stages: generation and alignment. In the generation stage, novel examples for a target class are created by utilizing the original data. \citet{lin2023selective} and \citet{sahu2022data} employ seed data as prompts to generate augmented examples. On the other hand, PromptMix \cite{sahu2023promptmix} and GPT3MIX \cite{Yoo2021} aim to enhance separability by incorporating data from two classes into their prompts. However, these approaches use a single fixed prompt and only target class seed data for generation. On the other hand, PromptMix \cite{sahu2023promptmix} and GPT3MIX \cite{Yoo2021} aim to enhance separability by incorporating both target and non-target class data. However, they either designate all classes or randomly sample classes as non-target without considering inter-class relevancy. These methods exhibit limitations in tasks involving numerous classes with significant inter-class relationship variances (e.g., intent classification).  There are two primary methods in the alignment stage: filtering and relabeling. Filtering \cite{lin2023selective} removes misaligned examples while relabeling \cite{sahu2022data,sahu2023promptmix} assigns new labels to examples based on classification results. However, these methods have not adequately addressed the inherent weaknesses caused by insufficient training data for verifying misaligned examples in few-shot settings. This can lead to FNs and an inability to handle OOD examples.

TARDiS addresses the limitations of existing methods by utilizing two generation processes, SEG and CEG, which leverage spark thoughts to create multiple class-specific prompts. Furthermore, it overcomes the limitations of existing alignment methods through CA.

\begin{figure*}[t]
    \centering
    \includegraphics[width=\textwidth]{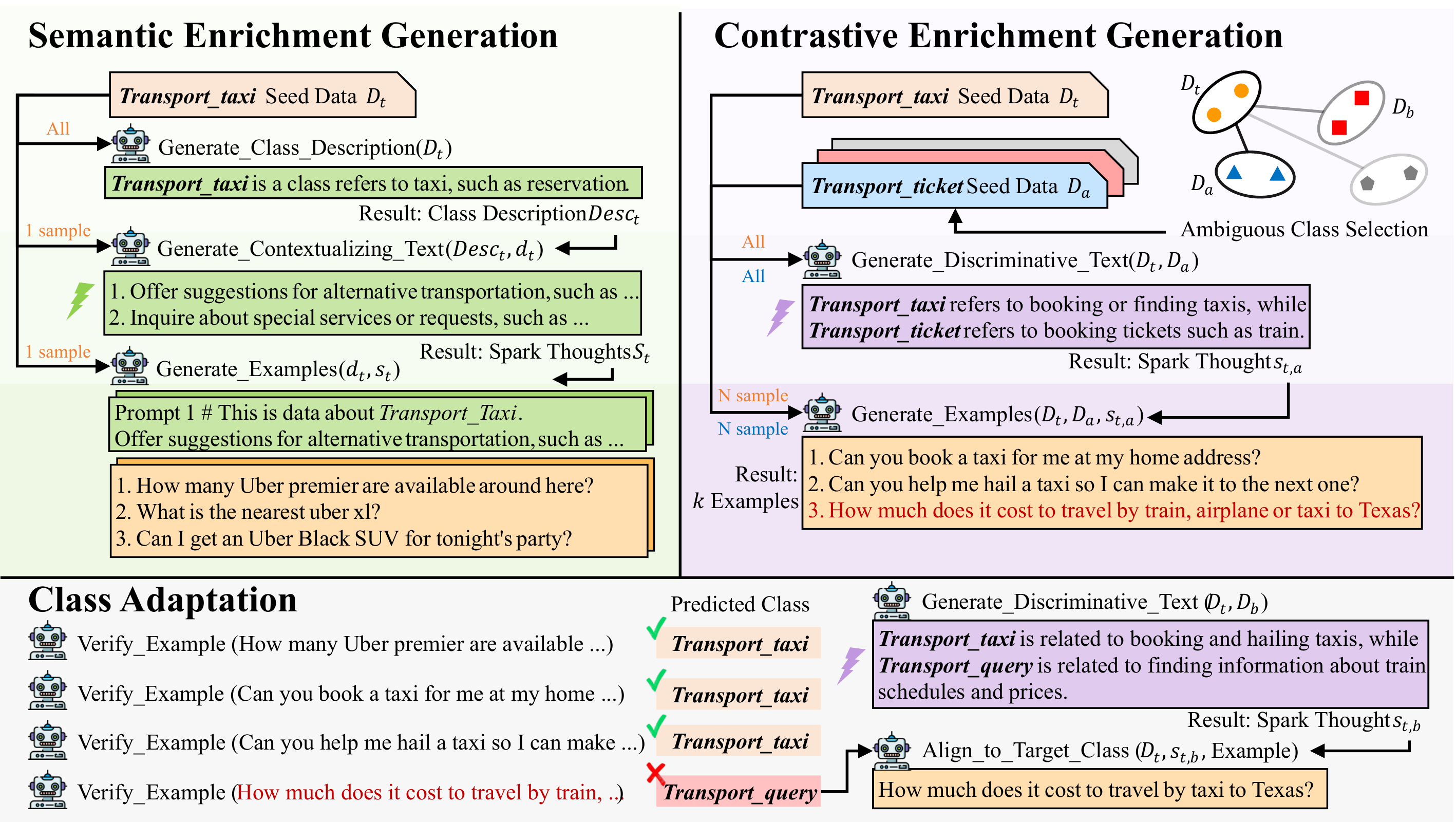}
    \caption{TARDiS framework. SEG generates contextualizing text using seed data, then generates target class example by utilizing contextualizing text. CEG generates discriminative text utilizing seed data from target and selected ambiguous class. CA verifies generated examples and modifies them to algin with the target class.}
    \label{fig:fig2}
\end{figure*}

\subsection{Chain of Thought Prompting}
LLMs such as GPT-3 \cite{Brown2020} and Llama \cite{Hugo2023}, possess high inferential and linguistic capabilities using a vast amount of internal knowledge \cite{Zhao2023}. 

CoT prompting is a method that maximizes the capabilities of LLMs \cite{Kojima2022,Wei2022}. It works by generating a flow of thoughts similar to human recognition, and reasoning to solve targeted tasks based on these. CoT prompting is successfully employed in various tasks such as logical reasoning \cite{ho-etal-2023-large} and question answering \cite{Lu2022,wang2023towards}.  Some research attempts to perform diverse and creative generation based on CoT prompting. Tree of thought methods \cite{Yao2023} generate various thoughts and select the most appropriate one for creative writing tasks. Attrprompt \cite{Yu2023} generates attributes and data based on them for synthetic data generation. However, these methods do not fully reflect the characteristics of TA tasks, such as generating diverse examples within a specific class while maintaining separability between classes. 

\section{Methodology}

TARDiS operates as in Figure~\ref{fig:fig2}. In the generation stage, two complementary methods, SEG and CEG, generate examples with high diversity and separability. Each method  utilizes seed data $D_t$ of a target class $t \in C$ to create class-specific spark thoughts $S_t$. $s_t$ and $D_t$ are combined to serve as multiple class-specific prompts for TA. In the alignment stage, a CA method ensures that the generated examples align with the target classes. Finally, these aligned examples are used as augmented data to train a classification model for evaluation.

\subsection{Generation Processes}

\subsection*{Semantic Enrichment Generation (SEG)}
LLMs tend to generate stereotypical and generic examples influenced by the common situations frequently occurring in extensive training data. By contextualizing input prompts with desired sentence styles, specific contexts, or particular attributes, LLMs can generate more diverse and contextually appropriate examples. SEG generates spark thoughts for class-specific conditioning to increase the diversity and relevancy of generated examples for each class.

The overall flow of SEG is illustrated in Figure~\ref{fig:fig2}. First, a class description $Desc_t$ is created using the seed data $D_t$. This description characterizes the target class, helping an LLM generate spark thoughts closely aligned with its semantic characteristics. For each seed example $d_t \in D_t$, multiple spark thoughts $s_t$ are generated using the class description $Desc_t$ and $d_t$, and are accumulated in $S_t$ without any post-processing. Finally, for each $s_t \in S_t$, $k$ examples are generated through $LLM(d_t, s_t)$.

\subsection*{Contrastive Enrichment Generation (CEG)}

SEG generates diverse examples using the seed data within the target class but does not guarantee separability between different classes among examples. In contrast, CEG performs inter-class conditioning using discriminative text as a spark thought to improve separability between non-target classes.

The overall flow of CEG is illustrated in Figure~\ref{fig:fig2}. First, unlike designating all classes or randomly sampling classes as non-target, we select ambiguous classes based on the class similarity of seed data. This approach assumes semantically similar data is more likely to be confused with the target class. The class similarity score between the target class $t$ and other class $c$ is calculated using Equation~\ref{eq:sim}:
\begin{equation}
Sim(t, c) = \frac{1}{|D_t| \times |D_c|} \sum_{d_t \in D_t} \sum_{d_c \in D_c} cos(d_t, d_c)
\label{eq:sim}
\end{equation}
where $cos(d_t, d_c)$ denotes the cosine similarity between two embeddings of examples $d_t$ and $d_c$, extracted using SBERT\footnote{We used 
sentence-transformers/all-mpnet-base-v2 for SBERT on sentence-transformers for every SBERT.}.
Then, for each class $t$, we select $n$ classes with the highest similarity scores as a set of ambiguous classes $A_t$. Subsequently, for $t$ and $a \in A_t$, $s_{t,a}$ is generated to state the differences between the two classes explicitly. Finally, $k$ examples are generated through $LLM(D_t, D_a, s_{t,a})$. To maximize the diversity of the input prompts, we randomly remove one or two examples from $D_t$ and $D_a$, varying the order of the remaining examples.

\subsection{Class Adaptation (CA)}

Misaligned examples are one of the primary factors that degrade the quality of augmented data. To address this issue, we propose a Class Adaptation (CA) method, which modifies misaligned examples to align with the target classes based on verification.  First, for each generated example, $m$ semantically similar examples are retrieved from the seed data based on the embeddings extracted by SBERT. These retrieved examples and their corresponding classes are used as shots to construct a verification prompt. The generated examples are verified using an LLM classifier with the verification prompt. If a prediction $p$ differs from $t$, it is considered misaligned. Each misaligned example $e$ is modified by $LLM(D_t, s_{t,p},e)$ to obtain $e'$ that aligns with class $t$, where $s_{t,p}$ is the same discriminative text used in CEG.

Our CA method offers two advantages. Firstly, CA can handle OOD classes. Even if a generated example is completely unrelated to the target domain or is not a proper sentence, it can be aligned with the target class through intensive modification. Secondly, CA can deal with false negatives induced by incorrect predictions from an LLM classifier. The meaning of a generated example can be preserved through minimal modification.

\section{Experimental Setup}

\begin{table}[t]
\centering
\begin{tabular}{lrrr}
\toprule
\textbf{Name}   & \textbf{\#Classes}    & \textbf{\#Train}  &\textbf{\#Test}    \\ \midrule
BANKING77       & 77                    & 8,632             & 3,084             \\
CLINC150        & 150                   & 15,000            & 4,500             \\
HWU64           & 64                    & 8,954             & 1,076             \\
TREC6           & 6                     & 5,453             & 500               \\
\bottomrule
\end{tabular}%
\label{tab:table1}
\caption{Data statistics. Official splits are used for TREC6 \cite{Li2002}, whereas other datasets are split by following DialoGLUE \cite{Mehri2020}.}
\end{table}

\subsection{Datasets and Evaluation }
We opted for four datasets widely used for few-shot classification: BANKING77, CLINC150, HWU64, and TREC6. BANKING77 \cite{Casanueva2020} focuses on the banking domain, while CLINC150 \cite{Larson2019} and HWU64 \cite{hwu64} cover 20 and 21 domains, respectively. TREC6 \cite{Li2002} aims at question classification in the open domain. Details of the datasets are provided in Table 1. For 5-shot and 10-shot settings on BANKING77, CLINC150, and HWU64, we followed the data split and seed data selection from DialoGLUE \cite{Mehri2020}. For the 2-shot setting on TREC6, we utilized the official splits and randomly sampled seed data from the training set.
As an evaluation metric, accuracy was selected for direct comparison with recent methods.

\subsection{Augmentation and Classification}
For augmentation with TARDiS, we used an instruction-tuned LLM, Llama2-13b \cite{Hugo2023}, was used. A repetition penalty of 1.15 \cite{Keskar2019} was applied to generate text different from the seed data. In 5-shot and 10-shot settings, augmentation was performed 50 times for both SEG and CEG, respectively, while 25 times in 2-shot settings.

For a fair comparison in classification, we finetuned and compared three PLMs, RoBERTa-base, RoBERTa-large \cite{Liu2019}, and BERT-base \cite{Devlin2019} as text classifiers trained with SOTA TA methods. We followed the standard classification procedure using an additional linear layer and “\textit{[CLS]}” token. To mitigate differences from a baseline due to the amount of training data, we finetuned PLM classifiers based on the training steps rather than the epochs.  

PLM classifiers were finetuned for 4,000 steps in 5-shot and 10-shot settings and 1,000 steps in 2-shot settings. The hyperparameters were set according to CPFT \cite{Zhang2021} in 5-shot and 10-shot settings and PromptMix in 2-shot settings. All experiments were conducted using 8 NVIDIA A6000 GPUs.

\begin{table*}[t]
\centering

\begin{tabularx}{\textwidth}{l *{6}{>{\centering\arraybackslash}X}}
\toprule
& \multicolumn{2}{c}{\textbf{BANKING77}} & \multicolumn{2}{c}{\textbf{CLINC150}} & \multicolumn{2}{c}{\textbf{HWU64}} \\
\cmidrule(lr){2-3} \cmidrule(lr){4-5} \cmidrule(lr){6-7}
\multicolumn{1}{c}{} &\multicolumn{1}{c}{5} & \multicolumn{1}{c}{10} & \multicolumn{1}{c}{5} & \multicolumn{1}{c}{10} & \multicolumn{1}{c}{5} & \multicolumn{1}{c}{10} \\
\midrule
\multicolumn{7}{c}{\texttt{RoBERTa-base}} \\
\midrule
baseline & 74.38 & 82.11 & 88.36 & 91.44 & 74.96 & 80.76 \\
conveRT & 75.32 & 83.32 & 89.22 & 92.62 & 76.95 & 82.65 \\
\citet{Mehri2021} & - & 85.95 & - & 93.97 & - & 86.28 \\
DNNC & 80.40 & 86.71 & 91.02 & 93.76 & 80.46 & 84.72 \\
CPFT & 80.86 & 87.20 & 92.34 & 94.18 & 82.03 & \textbf{87.13} \\
TARDiS & \textbf{83.50} & \textbf{87.30} & \textbf{93.34} & \textbf{94.96} & \textbf{83.42} & 86.52 \\
  w/o Multiple prompt &      81.27         &        86.33            &      92.49            &     94.09     &            82.16      &      83.92        \\
  w/o Multiple prompts \& CA & 80.38& 85.66 & 92.31 & 93.24 & 78.37 &81.35\\
\midrule
\multicolumn{7}{c}{\texttt{RoBERTa-large}} \\
\midrule
baseline & 80.32 & 86.52 & 91.11 & 92.95 & 79.18 & 84.57 \\
ICDA & 84.01 & \textbf{89.79} & 92.62 & 94.84 & 82.45 & 87.41 \\

TARDiS & \textbf{85.34} & 88.89 & \textbf{94.24} & \textbf{95.68} & \textbf{84.45} &\textbf{87.52} \\
  w/o Multiple prompts &           84.04    &              87.94      &  93.87                &      95.25    &          83.75        &            85.90  \\
  w/o Multiple prompts \& CA   & 80.95 &86.55 & 93.29  &                     94.57  &            81.29          & 84.67              \\
\bottomrule
\end{tabularx}
\caption{Average accuracies measured with three random seeds using RoBERTa-base (top) and RoBERTa-large (bottom) in 5-shot and 10-shot settings. We report the best results based on the quantities of data generated for ICDA. ``w/o Multiple prompts'' and ``w/o Multiple prompts \& CA'' indicate a single prompt with and without CA, respectively.}
\label{tab:table2}
\end{table*}

\subsection{Comparison Methods}
We selected seven recent TA methods as our comparison methods. PLMs trained with only few shots are regarded as baselines. \textbf{ConveRT} \cite{Henderson2020} is a dual-encoder intent detection model, pre-trained with pairs of input and response.   \textbf{\citet{Mehri2021}} is an intent detection model trained on CONVBERT \cite{Mehri2020} with the concept of attention observer and similarity matching. \textbf{DNNC} \cite{Zhang2020} is a discriminative model that identifies the most compatible example from a training set via similarity matching using RoBERTa-base model. \textbf{CPFT} \cite{Zhang2021} leverages contrastive learning to pretrain RoBERTa-base model on various intent classification datasets, followed by finetuning  on a specific target dataset. 
\textbf{ICDA} \cite{lin2023selective} is an LLM-based TA method that selects generated examples most helpful for training the model while filtering out others. \textbf{GPT3MIX} \cite{Yoo2021} integrates information between two classes through pseudo-labeling, aiming for separability. 
\textbf{PromptMix} \cite{sahu2023promptmix}, is LLM-based TA methods, which focuses on separability, fuses features across an LLM and manual descriptions. 

\section{Result}

\begin{table}
\centering
\begin{tabular}{lc}
\toprule
\textbf{Method}                 & \textbf{Accuracy} \\ \midrule
baseline                        & 40.3              \\
GPT3Mix(GPT3)                   & 57.3              \\
PromptMix(Llama-2-13b-chat-hf)  & 66.6              \\
PromptMix(Llama-2-70b-chat-hf)  & \textbf{70.8}     \\
TARDiS(Llama-2-13b-chat-hf)     & 70.2              \\
\bottomrule
\end{tabular}%
\caption{Average accuracies measured with three random seeds using BERT-base in 2-shot settings on TREC6. We report the best results from 1\% subsamples of TREC6 for GPT3MIX. An LLMs used for each method is denoted in parenthesis.}
\label{tab:table3}
\end{table}

\subsection{Main Performance}
Table~\ref{tab:table2}  presents the results of the comparison methods using RoBERTa-base (top) and RoBERTa-large (bottom) in both 5-shot and 10-shot settings on three datasets. TARDiS achieves significant improvements in most configurations, with two exceptions: CPFT with RoBERTa-base on HWU64 and ICDA with RoBERTa-large on BANKING77 in 10-shot settings. Despite these exceptions, TARDiS remains superior and competitive for two reasons. First, TARDiS operates without human-annotated data, whereas CPFT relies on various human-annotated data for intent classification. Second, TARDiS employs a relatively small LLM (i.e., Llama2-13b), whereas ICDA adopts OPT-66b \cite{Zhang2022}, which is about five times larger. Table~\ref{tab:table3} presents the performance in 2-shot settings. 

Furthermore, we analyzed the impact of each component in TARDiS. Compared to the complete TARDiS, the model without multiple prompts relies on a single fixed prompt, showing less effectiveness. Without multiple prompts and CA, the model cannot refine and align the generated examples with their target classes, leading to a significant performance drop. 

Table 3 shows that TARDiS outperforms the baseline, GPT3Mix, and PromptMix using Llama2-13b while demonstrating competitive results compared to PromptMix using Llama2-70b, which is about six times larger.   

\begin{table}[!t]
    \centering
    
\begin{tabular}{lccc}
\hline

\textbf{Method} & \textbf{Banking77} & \textbf{CLINC150} & \textbf{HWU64} \\ \hline
CEG only         & 84.36              & 93.13             & 83.27          \\
SEG only         & 84.56              & 92.75             & 82.34          \\
SEG + CEG        & \textbf{84.92  }            & \textbf{93.86 }            & \textbf{83.73  }        \\ \hline
\end{tabular}
    \caption{Performance comparison of generation methods on RoBERTa-large, banking77 5-shot setting. For each class, 250 examples are used.}
\label{tab:table6}
\end{table}

\subsection{Analysis}

\begin{table*}[!t]
    \centering
    \begin{minipage}{\textwidth}
    \begin{tabularx}{\textwidth}{@{}>{\arraybackslash}X>{\arraybackslash}X>{\arraybackslash}X@{}}
        \toprule Seed data of \textit{Transport\_taxi}  \\ \midrule
        - find transportation to rome &
        - look for a taxi to go home &
        - book a taxi at my home address for  \\
        - call me a cab to the train station &
        - taxi now  & right now\\
        \bottomrule
    \end{tabularx}
    \begin{center}
        (a) Seed data examples
    \end{center}
    
    \vspace{0.02em}

    \begin{tabularx}{\textwidth}{@{}p{0.3\linewidth}X@{}}
        \toprule
        Contextualizing text & Generated Samples - SEG\\
        \midrule
        Offer suggestions for alternative transportation options, such as public transit or ride-sharing services &
        - Could you compare costs for various taxi services like \textbf{Uber and Lyft}? \newline
        - Could I use \textbf{Uber} to get to the Pantheon rather than hailing a cab on the street?\newline
        - Should I book \textbf{regular or premium taxi}?\\
        \midrule
        Inquire about special services or requests, such as a specific type of vehicle or assistance for mobility issues. &
        - Can I get a taxi with a \textbf{handicap ramp}?\newline
        - Does this taxi support \textbf{infant car seats} because we are traveling with a newborn baby and need to ensure their safety during transit?\newline
        - I need a taxi that can provide \textbf{extra assistance for my elderly mother}\\
        \bottomrule
    \end{tabularx}
    \begin{center}
        (b) Contextualizing text and generated samples from SEG
    \end{center}
    
    \vspace{0.02em}

    \begin{tabularx}{\textwidth}{@{}p{0.3\linewidth}X@{}}
        \toprule
        Ambiguous class & Generated Samples - CEG\\
        \midrule
        \textit{Transport\_ticket} &
        - Hey, I need a ride to \textbf{JFK airport} ASAP! Can you call me a taxi?\newline
        - \textbf{Book a taxi} for an \textbf{airport} pickup right now!\newline
        - What's the estimated time of arrival for my \textbf{previously booked taxi}?\\
        \midrule
        \textit{Transport\_query} &
        - Please send me a taxi \textbf{as soon as possible; I'm stuck in the rain}!!\newline
        - I require an immediate pickup from my current location for a \textbf{quick drop-off} at the office.\newline
        - How about a \textbf{speedy transfer} to the hotel?\\
        \bottomrule
    \end{tabularx}
    \begin{center}
        (c) Ambiguous class and Generate samples from CEG
    \end{center}
\end{minipage}%
    \caption{Samples of generated examples through SEG and CEG for \textit{Transport\_taxi} class on HWU64. In SEG, features aligned with the contextualizing text are highlighted, while potentially confusable with ambiguous classes are highlighted in CEG.}
\label{tab:tablespark}
\end{table*}

\begin{table}[t]
\centering
{\small
\begin{tabular}{@{}lcccccc@{}}
    \toprule
             & \multicolumn{2}{c}{\textbf{BANKING77}}                       & \multicolumn{2}{c}{\textbf{HWU64}}                    & \multicolumn{2}{c}{\textbf{CLINC}}            \\\cmidrule(lr){2-3} \cmidrule(lr){4-5} \cmidrule(lr){6-7}
             & Inter                        & Intra                         & Inter                     & Intra                     & Inter                     & Intra             \\ \midrule
    5-shot   & 0.2138                       & 0.6662                        & 0.1001                    & 0.5262                    & 0.0905                    & 0.6285            \\
    Train    & 0.2191                       & \textbf{0.5531}               & 0.0952                    & 0.4161                    & 0.0898                    & 0.5369            \\
    AUG      & \textbf{0.2118}              & 0.5913                        & \textbf{0.0898}           & \textbf{0.4134}           & \textbf{0.0776}           & \textbf{0.4736}   \\ \bottomrule
\end{tabular}
}%
\caption{Inter-class and intra-class average pair similarity (APS) for different datasets and data subsets. Inter- and intra-class APS measures the average cosine similarity between different classes and within the same class, respectively.  Lower APS values indicate more diversity. The used embedding model is same with SBERT.}
\label{tab:table_aps}
\end{table}

\textbf{Measuring Diversity\quad} Table \ref{tab:table_aps} presents the Average Pair Similarity (APS) scores on three datasets, where a pair similarity is computed by the cosine similarity between two examples'  embeddings extracted by SBERT. We compute the inter-class and intra-class APS scores for 5-shot, training data (Train), and TARDiS's augmented data (AUG) for each dataset, respectively. It is assumed that lower intra-class APS values indicate high diversity within a class, whereas they are low diversity and vice versa. We achieve AUG\textless Train$\ll$5-shot on HWU64 and CLINC and Train\textless AUG$\ll$5-shot on BANKING77, respectively. Moreover, the inter-class APS scores remain similar, with minor changes across 5-shot, Train, and AUG. This indicates that TARDiS improves diversity over 5-shot at a minimum and ideally over Train while preserving the original distribution between classes.

Table \ref{tab:tablespark} shows samples of spark thoughts and generated examples using (b) SEG and (c) CEG from (a) the seed data for \textit{Transport\_taxi} class. In (b), SEG generates examples with new words, such as ``Lyft'' and ``handicap ramp'', following the intent of the spark thoughts to maximize diversity. In (c), CEG effectively generates examples by incorporating information from other classes to increase class separability. For instance, when \textit{Transport\_tickets}, associated with reservations for trains and airplanes, is used as the ambiguous class, references to these modes of transportation frequently appear in the generated examples. In contrast, when \textit{Transport\_query}, related to traffic situations, is employed, the generated examples primarily focus on traffic volume or time. \\

\noindent\textbf{Effect of SEG and CEG\quad}Table \ref{tab:table6} compares our generation methods. SEG and CEG show similar performance when used individually, varying their effectiveness across datasets. However, combining SEG and CEG consistently outperforms individual methods on three datasets. This indicates that each component plays a different role, and their combination implements a unique TA method.

\begin{figure}[t]
    \centering
    \includegraphics[width=1.0\linewidth]{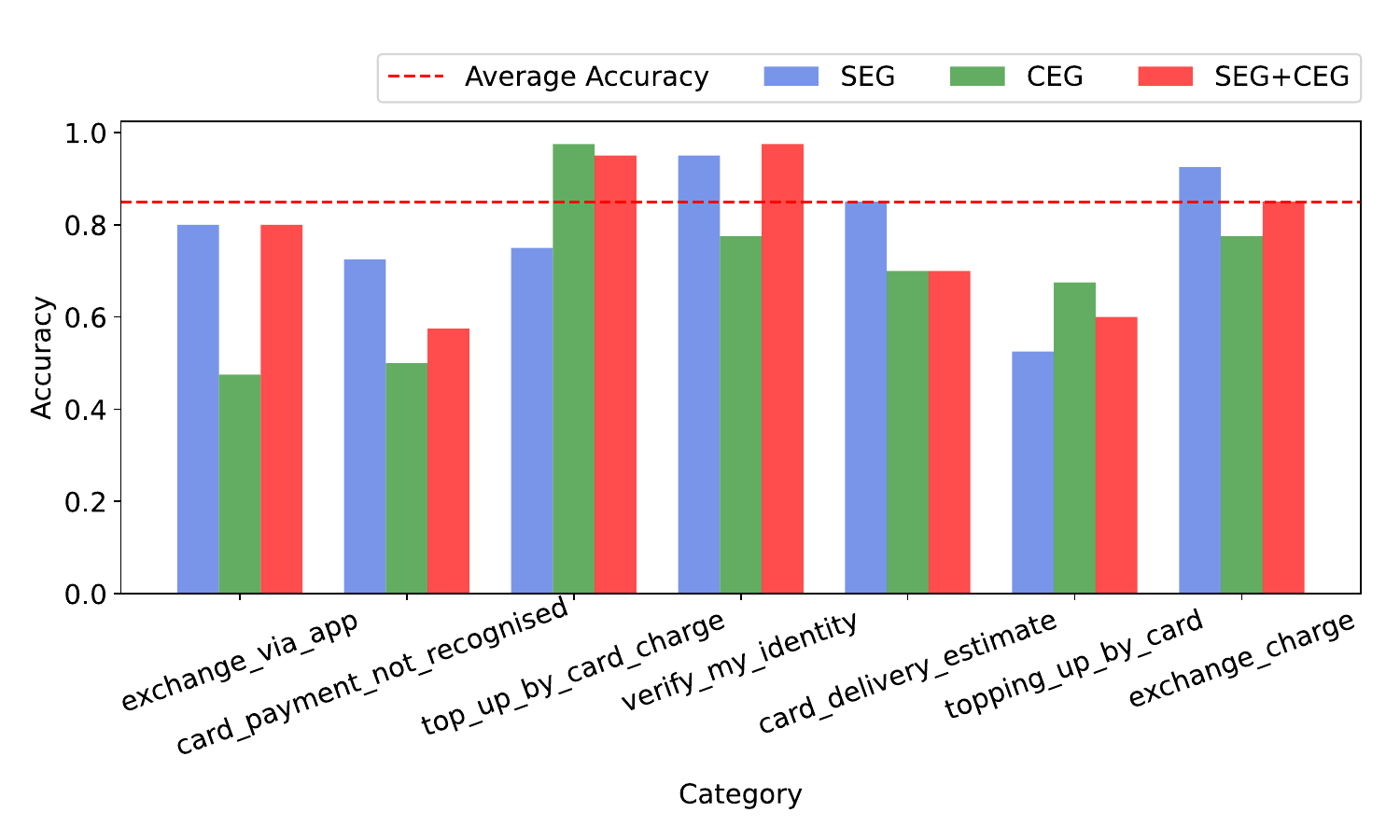}
    \caption{Class-wise performance comparison of SEG and CEG using RoBERTa-large in 5-shot settings on BANKING77. Results are sorted by the difference between SEG and CEG. The red line denotes an average  accuracy.}
    \label{fig:acc_comparison}
\end{figure}

Figure \ref{fig:acc_comparison} analyzes the effects of SEG and CEG on BANKING77. SEG and CEG exhibit different characteristics, with notable variations across specific classes. For instance, the \textit{Exchange\_via\_app} class encompasses a wide range of general exchanges, which contrasts with the discriminative text suggesting it should only include app-based exchanges. CEG overemphasizes the app aspect as the primary distinguishing factor from other classes, failing to capture the intrinsic diversity within the class. On the other hand, the low performance in the \textit{Top\_up\_by\_card\_charge} and \textit{Topping\_up\_by\_card} classes from SEG can be attributed to its insufficient consideration of class separability, failing to establish clear boundaries between these semantically related classes. Consequently, combining SEG and CEG improves diversity and separability in the resulting augmented data, leading to better performance over individual SEG and CEG.  

Table \ref{tab:table_nontargetselect} highlights the importance of selecting ambiguous classes in CEG. We achieve Top 1-5\textgreater Top 1-10\textgreater Top 6-10, indicating that focusing on strongly related classes is effective, whereas including less related classes is less effective because generated examples from selected classes may lack coherence and relevancy.   \\

\begin{table}[!t]
    \centering
    \begin{tabular}{l|c}
\hline
\textbf{Selected Ambiguous Class} & \textbf{Acc} \\
\hline
Random sampling & 81.33 \\
Top 1-10 & 82.43 \\
Top 6-10 & 81.13 \\
Top 1-5 & \textbf{83.18} \\
\hline
\end{tabular} %
    \caption{Performance comparison of non-target class selection with CEG using RoBERTa-base in 5-shot settings on Banking77. Top 1-5, 5-10, and 1-10 classes are selected by Equation 1 results. Random selection of 5 classes is included for comparison. SEG-generated examples are not used.}
\label{tab:table_nontargetselect}
\end{table}

\begin{figure}[]
    \centering
    \includegraphics[width=1.0\linewidth]{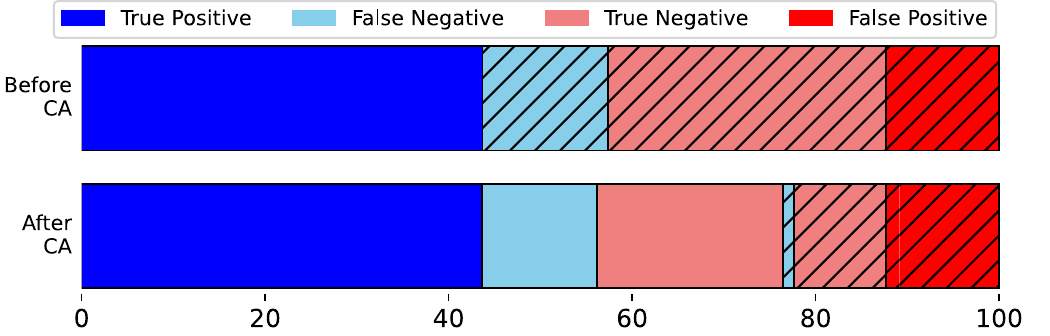}
    \caption{Two graphs of the proportional confusion matrices from the results of a verifier (i.e., an LLM classifier) in 5-shot settings on BANKING77, where the top and bottom denote before and after CA, respectively. The striped areas denote the proportions of misaligned (TN and FP) or potentially misaligned (FN). Ground truth labels, which serve as a basis for evaluating the performance of the LLM classifier, were obtained from a model trained on the entire dataset.}
    \label{fig:ca_anla}
\end{figure}

\noindent\textbf{Effects of Class Adaptation\quad} Figure \ref{fig:ca_anla} shows two graphs of the proportional confusion matrices before and after CA. At the top, a verifier misclassifies aligned examples as misaligned with a high proportion of 13\% and 12\% in FN and FP, respectively. A modifier only deals with FN and TN, which are classified as misaligned by the verifier. Consequently, our CA reduces the proportion of misaligned examples from 56\%, with the existing 43\% FP+TN and 13\% FN, to 23\%. Table \ref{tab:ca_example} demonstrates sample results applying CA, divided into (a) and (b) for TN and (c) for FN, respectively. For TN, (a) a part of an example and (b) an OOD example that is either irrelevant to the dataset or entirely incorrect are modified to align with the corresponding class. For FN, (c) an example is aligned to the target class with minimal modifications compared to filtering and relabeling, which can introduce additional misalignment. Despite those successes, FP cases remain a significant challenge, as they cannot be effectively handled using existing methods.

\begin{table}[t]
\centering
\small
\begin{tabularx}{\columnwidth}{lX}
\toprule
(a) & What's the quickest way \textbf{to get} to LaGuardia from here? 
\newline $\Rightarrow$ What's the fastest way \textbf{to hail a taxi} to LaGuardia from here? \\
\midrule
(b) & Where can I \textbf{~~~} 
\newline $\Rightarrow$ Where can I \textbf{find a reliable taxi service} \\
\midrule
(c) & How \textbf{long} will it take to reach Brooklyn Bridge from Times Square \textbf{by taxi?} 
\newline $\Rightarrow$ How \textbf{much} time will it take to get to Brooklyn Bridge from Times Square \textbf{by hailing a taxi?} \\
\bottomrule
\end{tabularx}%
\caption{Samples of before and after applying CA for \textit{Transport\_taxi} class on HWU64. The changes are highlighted.}
\label{tab:ca_example}
\end{table}

\section{Conclusion}
We propose TARDiS to overcome the limitations of existing two-stage LLM-based TA methods in few-shot settings. In the generation stage, SEG and CEG based on multiple class-specific prompts to enhance diversity and separability. In the alignment stage, CA method ensures that generated examples align with the corresponding target class, effectively dealing with examples that are misaligned, OOD, or FNs. We demonstrate the effectiveness of TARDiS by achieving SOTA performance on various few-shot text classification benchmarks and investigate detailed behaviors at each stage through an in-depth analysis. In future work, we plan to extend TARDiS to more challenging tasks such as creative writing and question answering.  Moreover, we plan to investigate efficient methods to reduce computational costs.

\section{limitations}

Although TARDiS demonstrates significant performance improvements through effective augmentation on various datasets, several limitations need to be addressed. First, even though the simultaneous use of SEG and CEG leads to better performance on average, we observe that a few classes have limited benefits. We hypothesize that this issue may arise when there are critical examples that can significantly hinder training. This study does not consider the characteristics and handling methods for such critical examples. Second, although CA effectively addresses the OOD and FN problems, it requires an extra process of regenerating all examples during the modification step. To mitigate this limitation, we will explore efficient modification methods focusing only on misaligned parts.

\bibliography{aaai25}

\section{Appendix}

\begin{table*}[t]{\fontsize{8pt}{9.6pt}\selectfont  
\resizebox{\textwidth}{!}{%
\begin{tabularx}{\textwidth}{@{}p{0.1\linewidth}X@{}}
            \toprule
            \textbf{Dataset} & \textbf{Prompt}\\
            \midrule
            BANKING77 & 
            \textbf{Generate\_Class\_Description (SEG)} : This is one of the classes in a intent classification dataset about banking \{target\_class\_name\} :\{target\_seed\_data\}. Describe this class in one sentence. 
            
            \par
            
            \textbf{Generate\_Contextualizing\_Texts (SEG)} : This is sentence from dataset for intent classification about banking. Considering the class, suggest a specific idea that can make the given dataset more diverse.\
The output format should summarize each idea in one sentence; example sentences are not required.
class name : \{data\}
class\_description : \{class\_description\} 
Sentence : \{target\_seed\_example\}  
            \par
            \textbf{Generate\_Examples (SEG)} : This is sentence from intent classification dataset about banking questions. Modify the given sentence to fit the characteristics presented. Give me five new modified texts. Class name : \{target\_class\} sentence : \{target\_seed\_example\} characteristics: \{contextualizing\_text\} 
            
            \par
            
            \textbf{Generate\_discriminative\_Texts (CEG)} : This is part of dataset for intent classification for intent classification.\{target\_class\_name\} :\{target\_seed\_data\}. \{ambiguous\_class\_name\} :\{ambiguous\_seed\_data\}.  
Tell me the main difference between \{ambiguous\_seed\_data\}   and \{Ambiguous\_class\_name\} in one sentence.

            \par
            
            \textbf{Generate\_Examples (CEG)} : \{target\_class\_name\} :\{target\_seed\_data\}. \{ambiguous\_class\_name\} :\{ambiguous\_seed\_data\}.  Discriminitive Text : \{discriminative\_text\}. This is intent classification dataset about banking. Based on the provided texts for each classes and the Discriminitive text highlighting their differences, Generate five new texts that could be confused with an \{ambiguous\_class\_name\ but clearly possess features that belong to the \{target\_class\_name\} \\

            \midrule

            HWU64 \par CLINC150 & 
            \textbf{Generate\_Class\_Description (SEG)} : This is one of the classes in a intent classification dataset about daily life \{target\_class\_name\} :\{target\_seed\_data\}. Describe this class in one sentence. 
            
            \par
            
            \textbf{Generate\_Contextualizing\_Texts (SEG)} : This is sentence from dataset for intent classification about daily life. Considering the class, suggest a specific idea that can make the given dataset more diverse.\
The output format should summarize each idea in one sentence; example sentences are not required.
class name : \{data\}
class\_description : \{class\_description\} 
Sentence : \{target\_seed\_example\}  
            \par
            \textbf{Generate\_Examples (SEG)} : This is sentence from intent classification dataset about daily life. Modify the given sentence to fit the characteristics presented. Give me five new modified texts. Class name : \{target\_class\} sentence : \{target\_seed\_example\} characteristics: \{contextualizing\_text\} 
            
            \par
            
            \textbf{Generate\_discriminative\_Texts (CEG)} : This is part of dataset for intent classification for intent classification.\{target\_class\_name\} :\{target\_seed\_data\}. \{ambiguous\_class\_name\} :\{ambiguous\_seed\_data\}.  
Tell me the main difference between \{ambiguous\_seed\_data\}   and \{Ambiguous\_class\_name\} in one sentence.

            \par
            
            \textbf{Generate\_Examples (CEG)} : \{target\_class\_name\} :\{target\_seed\_data\}. \{ambiguous\_class\_name\} :\{ambiguous\_seed\_data\}.  Discriminitive Text : \{discriminative\_text\}. This is intent classification dataset about daily life. Based on the provided texts for each classes and the Discriminitive text highlighting their differences, Generate five new texts that could be confused with an \{ambiguous\_class\_name\ but clearly possess features that belong to the \{target\_class\_name\} \\

            \midrule
            TREC6 & 
            \textbf{Generate\_Class\_Description (SEG)} : This is one of the classes in a classification dataset about question type. \{target\_class\_name\} :\{target\_seed\_data\}. Describe this class in one sentence.
            
            \par
            
            \textbf{Generate\_Contextualizing\_Texts (SEG)} : This is sentence from dataset for question type classification. Considering the class, suggest five ideas that can make the given class more diverse.\
The output format should summarize each idea in one sentence; example sentences are not required.
class name : \{data\}
class\_description : \{class\_description\} 
Sentence : \{target\_seed\_example\}  
            \par
            \textbf{Generate\_Examples (SEG)} : This is sentence from dataset about question type classification. Reference the given sentence and generate new data for  Class name : \{target\_class\} sentence : \{target\_seed\_example\} characteristics: \{contextualizing\_text\} 
            
            \par
            
            \textbf{Generate\_discriminative\_Texts (CEG)} :This is part of dataset for classfication. Each class have different answer type.\{target\_class\_name\} :\{target\_seed\_data\}. \{ambiguous\_class\_name\} :\{ambiguous\_seed\_data\}.  
Focus on the answer type, and tell me the main difference between \{ambiguous\_seed\_data\}   and \{Ambiguous\_class\_name\} in one sentence.

            \par
            
            \textbf{Generate\_Examples (CEG)} : \{target\_class\_name\} :\{target\_seed\_data\}. \{ambiguous\_class\_name\} :\{ambiguous\_seed\_data\}.  Discriminitive Text : \{discriminative\_text\}. This is classification dataset about question type. Based on the provided texts for each classes and the distinctive text highlighting their differences, generate five new texts that emphasize the unique characteristics of class {target\_class}. 
Generate texts for fit in  \{target\_class\_name\}  class. \\
            \bottomrule
\end{tabularx}
}}
\caption{Generation process prompts for different datasets.}
\label{tab:APPENDIX1-1}
\end{table*}

\begin{table*}[t]
\centering

\fontsize{8pt}{9.6pt}\selectfont
\resizebox{\textwidth}{!}{%
\begin{tabularx}{\textwidth}{@{}X@{}}
\toprule
\textbf{Prompt} \\
\midrule
\textbf{Verification}:

text: \{similar\_text\_1\} class: \{class\_1\} \\
text: \{similar\_text\_2\} class: \{class\_2\} \\
text: \{similar\_text\_n\} class: \{class\_n\} \\
text: \{target\_text\} class: \\[1ex]

\textbf{Modification}: \{target\_class\}: \{target\_class\_data\} \\
Discriminative text: \{discriminative\_text\} \\
This is query text which belongs to class \{verification\_result\_class\}. \\
Query text: '\{generated\_example\}' \\
Modify this query text to be suitable for \{target\_class\}. \\
\bottomrule
\end{tabularx}%
}

\caption{CA prompts for every datasets.}
\label{tab:APPENDIX2}
\end{table*}

\label{sec:appendix}
\subsection{The Prompt Formats}
Table \ref{tab:APPENDIX1-1},\ref{tab:APPENDIX2} shows our prompts used for different datasets. Prompts are applied in the bolded parts. Texts such as class names or examples inside the curly braces are inserted.

Continued on the next page.

\end{document}